\documentclass{article}
\PassOptionsToPackage{numbers, compress}{natbib}

\usepackage[final]{neurips_2019}
\usepackage{booktabs}
\usepackage{multirow}
\usepackage{caption}
\usepackage[dvipsnames]{xcolor}

\usepackage[utf8]{inputenc} 
\usepackage[T1]{fontenc}    
\usepackage{hyperref}       
\usepackage{url}            
\usepackage{booktabs}       
\usepackage{amsfonts}       
\usepackage{nicefrac}       
\usepackage{microtype}      
\usepackage{ bbold }
\usepackage{amsmath}
\usepackage{wrapfig}
\usepackage{pifont}
\newcommand{\cmark}{\ding{51}}%
\newcommand{\xmark}{\ding{55}}%
\usepackage{algorithmic}
\usepackage{algorithm}

\newcommand{\comment}[1]{}
\usepackage{graphicx}
\graphicspath{ {./Images} }

\title{Learning Soft Labels via Meta Learning}

\author{Nidhi Vyas \\Apple \\ \texttt{nidhi\_vyas@apple.com} \And Shreyas Saxena \\Apple\\ \texttt{shreyas\_saxena@apple.com} \And Thomas Voice \\Apple \\ \texttt{tvoice@apple.com} }

\begin{document}

\maketitle
\begin{abstract}
One-hot labels do not represent soft decision boundaries among concepts, and hence, models trained on them are prone to overfitting. Using soft labels \cite{szegedy2015rethinking} as targets provide regularization, but different soft labels might be optimal at different stages of optimization.
Also, training with fixed labels in the presence of noisy annotations leads to worse generalization. 
To address these limitations, we propose a framework, where we treat the labels as learnable parameters, and optimize them along with model parameters. 
The learned labels continuously adapt themselves to the model’s state, thereby providing dynamic regularization. 

When applied to the task of supervised image-classification, our method leads to consistent gains across different datasets and architectures.
For instance, dynamically learned labels improve ResNet18 by 2.1\% on CIFAR100. 
When applied to dataset containing noisy labels, the learned labels correct the annotation mistakes, and improves over state-of-the-art by a significant margin. 
Finally, we show that learned labels capture semantic relationship between classes, and thereby improve teacher models for the downstream task of distillation.

\end{abstract}
\section{Introduction}

In a general setting, training machine learning models for classification involves optimizing a mapping between input and fixed one-hot labels. 
 
Having fixed one-hot labels during training can be sub-optimal for multiple reasons. First, one-hot labels do not reflect the similarities between semantically related classes. Therefore, training models using them can lead to hard decision boundaries that do not generalize well
(over-fitting). Second, the annotated labels might contain noise, training on which can degrade performance \cite{han2018co}. Various techniques have been proposed to tackle these problems separately. For example, in label smoothing \cite{Rafael2019, szegedy2015rethinking}, the one-hot labels are converted into soft labels by uniformly distributing a small amount of probability mass on non-target classes, thereby providing regularization for learned decision boundaries. In \cite{shu2019metaweightnet}, an auxiliary neural network is used to assign weights to inputs, where the method ignores the noisy samples by assigning smaller weights to them.

\begin{figure}[!htbp]
\centering
\begin{minipage}[b]{0.35\textwidth}
  \centering
    \includegraphics[scale=0.32]{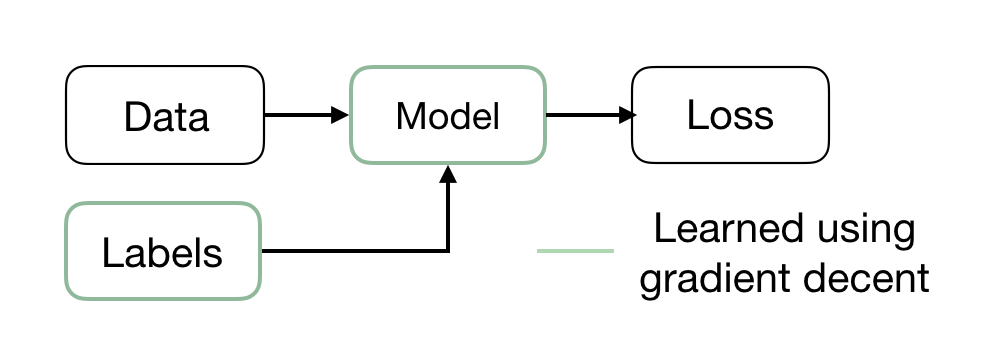}
\end{minipage}%
\hfill
\begin{minipage}[b]{0.64\textwidth}
  \centering
    \includegraphics[scale=0.23]{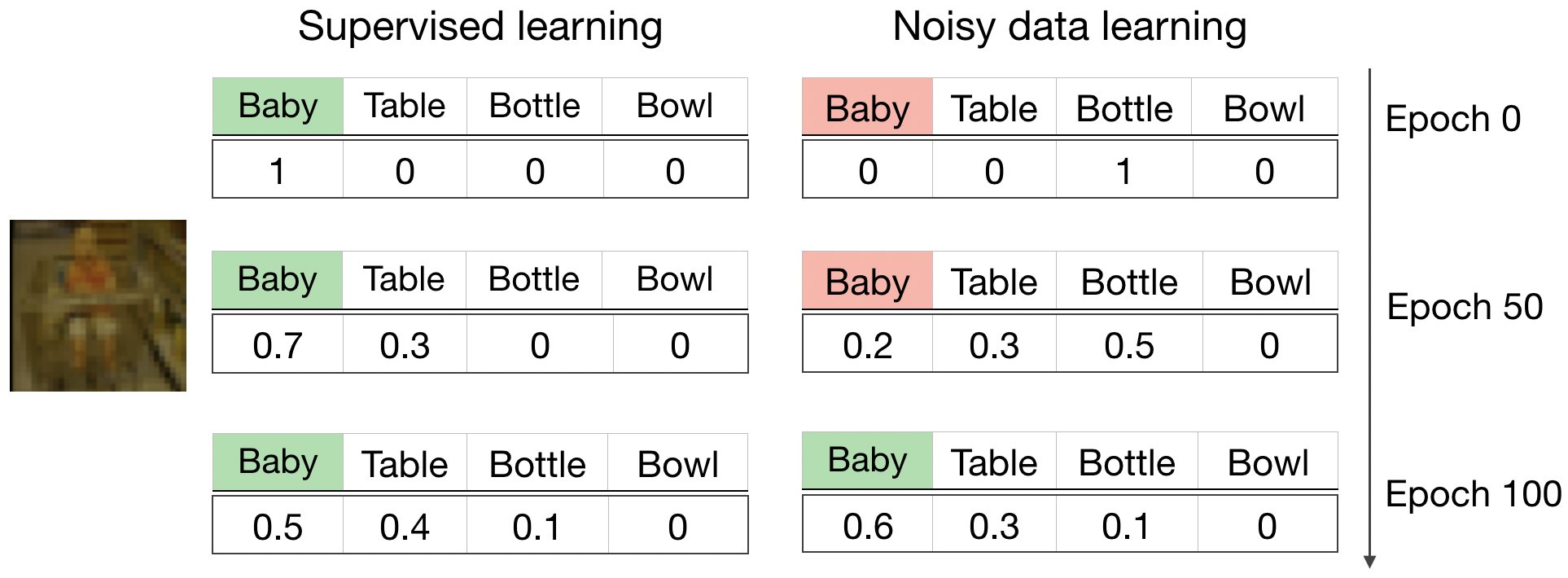}
\end{minipage}%
\caption{\textbf{Left}: Our method treats the labels as learnable parameters, and learns them along with the model parameters. \textbf{Right}: In supervised setting, our method starts with one-hot labels, and captures shared concepts specific to the instance. In noisy data setting, our method learns to correct the labels.}
\label{fig:n_c_overview}
\end{figure}

Obtaining the right labelling of a dataset to address the two problems mentioned above is a hard task \cite{LalorWY17}. One can generate soft labels using distillation \cite{hinton2015distilling}, but the labels obtained at convergence might not be optimal for all stages of training. Similarly, in settings where training data may contain wrong annotations, the learning process should ideally correct them, without relying on human intervention. In our work, we address these concerns by proposing an optimization framework common to both, supervised and noisy data settings. Under our framework, the labels are treated as learnable parameters, and are learned along with the model parameters  (see Figure \ref{fig:n_c_overview}). More specifically, throughout training, the framework looks at the performance of the model on a held-out set, and updates the labels on the training set, such that when the model is trained with these labels, it performs well on the held-out set. As a result, the learned labels are dynamic and adapt to different stages of training, leading to improved generalization. 
The main contributions of our work are:

\begin{enumerate}
\item We introduce a meta-learning framework where the labels of training set are treated as learnable parameters and are learned along with the model parameters. Labels are learned using gradient descent, and continously adapt to the model state during optimization.

\item When applied in supervised setting, training with dynamically learned labels improves over competitive baselines on CIFAR100 and CIFAR10 dataset across multiple architectures. 

\item We show that learned labels capture semantic relationship among classes, and training with them leads to better teacher models for the downstream task of distillation. Further, we show that these labels provide a competitive alternative to knowledge distillation, as they are directly transferable to other architectures. 

\item When used in noisy data setting, our framework corrects annotations mistakes in the dataset, and outperforms the state-of-the-art by significant margin. 
\end{enumerate}

\section{Learning Dynamic Soft Labels via Meta Learning}

As mentioned earlier, in contrast to using fixed {one-hot} labels for model optimization, our method learns dynamic soft labels that adapt to model state during training. In this section, we formalize our method and present the framework by first {showing how to learn} {class} labels, which are unique to every class in the dataset (see Section \ref{sec:class-specific}). Next, we show how our method can be extended to learn  {instance} labels, which are unique to every instance in the dataset (see Section \ref{sec:instance-specific}).

\subsection{Class labels}
\label{sec:class-specific}

At different steps of optimization, certain classes may be more prone to over-fitting than others. So, different degrees of regularization may be required for them. We begin by considering an adaptive smoothing parameter unique to each class, such that the class label is obtained by smoothing the one-hot labels proportional to this parameter. The parameters regularize each class differently, thereby preventing over-confident model predictions \cite{pereyra2017regularizing}.

To formalize, given a training set with \textit{N} data points and \textit{c} classes, let $\mathbf{x}_i \in \mathbb{R}^d$ 
denote a single data point and $y_i \in \{1,...,c\}$ denote the corresponding target class. Let ${\alpha}^t \in \mathbb{R}^c$ represent smoothing parameters for all classes at step $t$ of optimization, where $\alpha_k^t$ is the smoothing parameter for class $k$. Let ${p}_{i,k}^t$ be the probability mass on class \textit{k} for instance \textit{i} at step $t$. The effective class label $\mathbf{p}_i^t \in \mathbb{R}^c$ is a probability distribution defined as follows:

\vspace{-0.25cm}
\[
{p}_{i,k}^t (\alpha_{y_i}^t)= 
\begin{cases}
    1-\alpha_{y_i}^t,& \text{if } k = y_i\\
    \alpha_{y_i}^t / (c-1),              & \text{otherwise}
\end{cases}
\]

Let, $f(\mathbf{x}, \theta^t)$ and $\theta^t$ denote the model and  model's parameters at step $t$ respectively.
Let $L_{train}(\mathbf{x}_i, \mathbf{p}_i^t; \theta^t)$ denote the cross-entropy loss between model prediction and class label $\mathbf{p}_i^t$ for data point $i$. For sake of explanation, we denote $L_{train}(\mathbf{x}_i, \mathbf{p}_i^t; \theta^t)$ as $\mathcal{L}_{i}^\textrm{train}({\alpha}^{t}_{y_i} ;\theta^t)$, as $\mathbf{p}_i^t$ is a function of ${\alpha}^t_{y_i}$. Our goal is to solve for optimal $\hat{\theta}$ by minimizing the loss on the training set.
\begin{align}
\mathcal{L}_{train}(\theta^t,\alpha^{t}) &= \frac{1}{N} \sum_{i=1}^N  \mathcal{L}_{i}^\textrm{train}(\alpha^t_{y_i} ;\theta^t)
\label{eq:sgd_update} \\
\hat{\theta}(\alpha^{t}) &= \arg \min_{\theta} \mathcal{L}_{train}(\theta, {\alpha}^{t})
\label{eq:loss_min}
\end{align}

\vspace{-0.25cm}
Here,  $\hat{\theta}$ is a function of $\alpha^t$. Note, if $\forall{t}$ $\alpha^t=0$, Eq. \ref{eq:sgd_update} is equivalent to standard classification loss with one-hot fixed targets. But, here, {the best choice of} $\alpha^{t}$ is not known \textit{apriori}. We cannot optimize $\alpha^t$ and $\theta^t$ {using the loss objective} on the same data points, because it can lead to degeneracy. For instance, Eq. \ref{eq:loss_min} allows for a local minima where the class labels and model predictions converge to each other and can be away from the ground truth. To avoid this issue, we turn to meta-learning, where we use a training set to optimize $\theta^t$, and a meta set to optimize $\alpha^t$. To save the computational cost of unrolling the feed-forward graph through multiple time steps, we use a one-step look ahead approach to estimate the gradient of $\alpha^t$ on meta set.

\textbf{Meta learning for class labels}:
We sample a mini-batch with $n$ instances from the training set, and simulate a one step look-ahead SGD update on model parameters (${\theta^{t+1}}^*$) as a function of $\alpha^{t}$ (see Eq. \ref{eq:one_step_sgd_roll_out_instance}). Here, $\lambda_{\theta}$ is the learning rate for model parameters. Next, the one step look-ahead update is used to compute the loss on meta set $\mathcal{L}^\textrm{meta}({\theta^{t+1}}^*)$. In our experiments, we use the cross entropy loss between the model predictions and labels in the meta set. This loss is used to compute the meta-gradient on the smoothing parameters (see Eq. \ref{eq:meta_gradient_instance}):

\vspace{-0.25cm}
\begin{align}
{\theta^{t+1}}^* (\alpha^{t}) &= 
\theta^{t} - \frac{\lambda_{\theta}}{n} \sum_{i=1}^n \frac{\partial \mathcal{L}_i^\textrm{train}(\theta^t, \alpha^{t}_{y_i})}{\partial \theta^{t}}
\label{eq:one_step_sgd_roll_out_instance} \\
\frac{\partial \mathcal{L}^\textrm{meta}({\theta^{t+1}}^*)}{\partial \alpha^{t}_{y_i}} &= 
\frac{\partial \mathcal{L}^\textrm{meta}({\theta^{t+1}}^*)}{\partial {\theta^{t+1}}^*} 
\cdot \frac{\partial {\theta^{t+1}}^*}{\partial \alpha^{t}_{y_i}} \nonumber \\
\frac{\partial \mathcal{L}^\textrm{meta}({\theta^{t+1}}^*)}{\partial \alpha^{t}_{y_i}} &=  
\frac{-\lambda_\theta}{n} \cdot \frac{\partial \mathcal{L}^\textrm{meta}({\theta^{t+1}}^*)} {\partial {\theta^{t+1}}^*} \cdot 
\frac{\partial}{\partial \alpha^t_{y_i}}\big[ {\frac{\partial \mathcal{L}^\textrm{train}_i}{\partial \theta^{t}}} \big]^T \label{eq:meta_gradient_instance}
\end{align}

Using the meta-gradient (in Eq. \ref{eq:meta_gradient_instance}), we update the smoothing parameter for each class using the first order gradient update rule (see Eq. \ref{eq:t_plus_1_update_inst_params}). The updated value $\alpha^{t+1}$ is then used to update the model parameters (see Eq. \ref{eq:t_plus_1_update_model}). Here, $\lambda_{\alpha}$ is the learning rate for smoothing parameters of all classes. 
\begin{align}
\alpha^{t+1}_{y_i} &= \alpha^{t}_{y_i} - \lambda_{\alpha} \frac{\partial \mathcal{L}^\textrm{meta}({\theta^{t+1}}^*)}{\partial \alpha^{t}_{y_i}}
\label{eq:t_plus_1_update_inst_params} \\
\theta^{t+1} &= 
\theta^{t} - \frac{\lambda_{\theta}}{n} \sum_{i=1}^n \frac{\partial \mathcal{L}_i^\textrm{train}(\theta^t, \alpha^{t+1}_{y_i})}{\partial \theta^{t}}
\label{eq:t_plus_1_update_model}
\end{align}

Our method is summarized in Algorithm \ref{alg}. Note, if the smoothing parameters are fixed and common to all classes, our method is similar to label smoothing \cite{szegedy2015rethinking}. However, having fixed smoothing throughout training might be suboptimal. We later verify this finding in Section \ref{sec:experiments-analysis}, where we show that when we adapt these parameters dynamically, the model prefers a higher degree of smoothing at the start of optimization, but reduces the smoothing at convergence. 

\begin{algorithm}[!htbp]
\vspace{0mm}
\renewcommand{\algorithmicrequire}{\textbf{Input:}}
\renewcommand{\algorithmicensure}{\textbf{Output:}}
\caption{Algorithm for learning dynamic soft labels via meta learning}
\label{alg:example}
\begin{algorithmic}[1]  \small
	\REQUIRE Train set $\mathcal{D}^\textrm{train}$, meta set $\mathcal{D}^\textrm{meta}$, learning rate of model $\lambda_\theta$, learning rate of class smoothing parameters $\lambda_{\alpha}$, max iterations $T$.
	\ENSURE  Model parameters at convergence $\theta^{T}$ 
	\STATE Initialize model parameters $\theta^{0}$ with random and $\alpha^0$ with 0.
	\FOR{$t=0$ {\bfseries to} $T-1$}
	\STATE $\{x^\textrm{train}, y^\textrm{train}\} \leftarrow$ SampleMiniBatch($\mathcal{D}^\textrm{train}$).
	\STATE $\{x^\textrm{meta},y^\textrm{meta}\} \leftarrow$ SampleMiniBatch($\mathcal{D}^\textrm{meta}$).
	\STATE $\{\mathbf{p}^{t}\} \leftarrow$ CalculateSoftLabels($\alpha^{t}$, $y^\textrm{train}$).
	\STATE Compute one step update for model parameters as function of $\alpha^{t}$ by Eq. 
        (\ref{eq:one_step_sgd_roll_out_instance}).
	\STATE Update $\alpha^{t+1}$ by Eq. (\ref{eq:t_plus_1_update_inst_params}).
	\STATE Update $\theta^{t+1}$ by Eq. (\ref{eq:t_plus_1_update_model}).
	\ENDFOR
\end{algorithmic}
\label{alg}
\end{algorithm}

\subsection{Instance labels}
\label{sec:instance-specific}
Class labels assume a homogeneous distribution of instances within a class. Further, they allocate uniform probability on non-target classes, meaning for a given class, all non-target classes are equally likely. However, instances within a class can have varying properties. For example, in Figure \ref{fig:n_c_overview} (right), consider an image of class \textit{Baby}, where the baby is sitting on a table. The ideal soft label should have more probability mass on class \textit{Table} than other non-target classes. Such instance labels {could} help learn softer decision boundaries via shared concepts. We use our meta-learning framework to learn such labels directly.

Formally, for a given instance $\mathbf{x}_i$ at step $t$ of optimization, instance labels are learnt via smoothing parameters, where $\alpha_{i,k}^t$ is the smoothing parameter for hypothesis class $k \in \{1,...,c\}$. The instance label $\mathbf{p}_i^t \in \mathbb{R}^c$, is a probability distribution over all $c$ classes, such that $\sum_{k=1}^c \alpha_{i,k}^t=1$. Here, ${p}_{i,k}^t (\alpha_{i,k}^t) = \alpha_{i,k}^t$
as shown in Eq. \ref{eq:rewrite_meta_gradient_update}, the meta gradient on an instance is proportional to the dot product of the gradient of training loss at the $i^{th}$ sample at step $t$, and the gradient of the meta loss over all samples in the meta set at step $t+1$. Therefore, a hypothesis class whose gradient aligns with the gradients on the meta set will obtain a higher probability mass. This allows a more informed allocation of probability mass on non target classes. 
\vspace{-0.25cm}

\begin{align}
\frac{\partial \mathcal{L}^\textrm{meta}({\theta^{t+1}}^*)}{\partial \alpha_{i,k}^{t}} &=  
\frac{-\lambda_\theta}{n} \cdot \frac{\partial \mathcal{L}^\textrm{meta}({\theta^{t+1}}^*)} {\partial {\theta^{t+1}}^*} \cdot 
\frac{\partial}{\partial \alpha_{i,k}^t}\big[ {\frac{\partial \mathcal{L}^\textrm{train}_i}{\partial \theta^{t}}} \big]^T \label{eq:rewrite_meta_gradient_update}
\vspace{-0.25cm}
\end{align}

Another advantage of learning instance labels is in the noisy data setting, as our framework can perform label correction. More specifically, the incorrect hypothesis class will misalign with the meta-gradients in Eq. \ref{eq:rewrite_meta_gradient_update}. So, under our algorithm, the probability mass over incorrect class will reduce, and the probability mass on correct class will increase during optimization (see Section \ref{sec:noisy}).

\section{Experimental Evaluation}
\label{sec:experiments}

\subsection{Implementation details}

To show the general applicability of our method, we evaluate our method on the CIFAR100 and CIFAR10 datasets \cite{Krizhevsky09learningmultiple} using multiple architectures viz. WideResNet (WRN) \cite{BMVC2016_87}, VGG16 \cite{simonyan2014very}, ResNet18 \cite{he2016deep}, and AlexNet \cite{krizhevsky2012imagenet}. Both datasets contain 50,000 images in the training set, and 10,000 images in the test set. Each image in CIFAR100 belongs to one of 100 classes, and each image in CIFAR10 belongs to one of 10 classes. We use 20\% of the training set as common held-out data for both validation set (tune hyper parameters) and meta set. The class and instance labels are optimized using a \textit{5}-fold cross validation approach. The final performance is reported on the full training data by averaging the learned dynamic labels across each fold. The label parameters do not change the overhead at inference, as they are only present during training. 

Unless stated otherwise, the following hyper-parameters are used for our meta-learning approach. Batch size for training and meta set is fixed to 128 and 256. We use separate optimizers for learning soft labels and model parameters, and both are optimized using stochastic gradient decent (SGD).
For class labels, we ensure a valid probability distribution by clamping the smoothing parameters between 0 and 1. For instance labels, we use the softmax function to ensure a probability distribution over all classes. We search in the range of \{0.3, 0.5, 0.7, 0.9\} for initializing the target classes for soft label parameters. The remaining probability mass is used to uniformly initialize the non target classes. We search in the range of \{25, 50, 75, 100, 150, 200\} for the learning rate of soft label parameters. These parameters do not use any learning rate schedule. For all experiments, we report the mean and standard deviation over 3 runs. 

\subsection{Image classification in supervised setting}

In this section we demonstrate the efficacy of our method when applied to the task of image classification in the supervised setting. We evaluate our framework on CIFAR100 and CIFAR10 dataset using ResNet18 \cite{he2016deep} and VGG16 \cite{simonyan2014very} architectures, and compare it with baseline which uses fixed one-hot labels. As shown in Table \ref{tab:baseline_results}, learning class labels on CIFAR100 improves over baseline by 1.8\% and 0.4\% for ResNet18 and VGG16, respectively. 
Learning instance labels lead to a gain of 2.1\% and 0.8\% for ResNet18 and VGG16, respectively. This result supports our hypothesis: instance labels relax the uniform prior over non-target classes, giving it an advantage to account for instance specific features. The gains over baseline are higher in ResNet18 compared to VGG16. We attribute this to ResNet18 being more prone to overfitting, and thereby benefiting more from dynamic regularization of soft labels. On CIFAR10, similar to CIFAR100, we improve over baseline. However, there are fewer, but more distinct classes, and so the gains are diminished. 

\begin{table}[!htbp]
\small
\centering
\begin{tabular}{@{}lll|ll@{}}
\toprule
                     & \multicolumn{2}{c|}{{CIFAR100}}                              & \multicolumn{2}{c}{CIFAR10}                                  \\ 
\multicolumn{1}{c}{} & \multicolumn{1}{c}{ResNet18} & \multicolumn{1}{c|}{VGG16} & \multicolumn{1}{c}{ResNet18} & \multicolumn{1}{c}{VGG16} \\ \midrule
Baseline             &             77.1 $\pm$ 0.3 \comment{99q3tvy7m5 }                 &    74.5 $\pm$ 0.2 \comment{bsuahv3q2j}                       &       95.0 $\pm$ 0.1 \comment{y85pqbdatd} &    93.9 $\pm$ 0.0 \comment{h6vt8f84ka}                               \\ \midrule
Ours (Class)               &  78.8 $\pm$ 0.2  \comment{jpw74pwrtn hqirpz5w3e 56py2frtgz } & 74.9  $\pm$ 0.2 \comment{ymwcgavkr9 dvakzbp4y5 fjnfxsrd5s}  & 95.2 $\pm$ 0.1 \comment{z6evrs493t zec43tsuj9 8by6q24ga8} & 94.1 $\pm$ 0.1 \comment{adu6wntpt4 4y88ikyn9n bzgd33nnys} \\
Ours (Instance)            &  \textbf{79.2 $\pm$ 0.2} \comment{S2 7zww5zc8bi S3 c9w2rags7c S1 ifxp44k6sq} & \textbf{75.3 $\pm$ 0.3} \comment{3acmxmvd8b S2 d5dgdh9fbb S1 igfz37r7xq} & \textbf{95.3 $\pm$ 0.1}  \comment{mvxfv33ij4 gqkac9zqa7 qb3q3qepxj} & \textbf{94.3 $\pm$ 0.1} \comment{hkdpb7fftk gzd442byam v84hf6e6df} \\ \bottomrule
\end{tabular}
\caption{Our method of learning dynamic labels achieves consistent gains over baseline.}
\label{tab:baseline_results}
\end{table}

\begin{wrapfigure}{r}{0.40\columnwidth}
  \centering
    \resizebox{0.40\columnwidth}{!}{
    \begin{tabular}{@{}lll@{}} \toprule
                   & ResNet18 & VGG16 \\ \midrule
        Baseline           &    77.1 $\pm$ 0.3 \comment{99q3tvy7m5 } &  74.5 $\pm$ 0.2 \comment{bsuahv3q2j}    \\
        Confidence Penalty \cite{pereyra2017regularizing} &       77.6 $\pm$ 0.3 \comment{yjudze937b}  & \comment{e4ypv7zkdu} 74.1 $\pm$ 0.1      \\ 
        Label Smoothing \cite{szegedy2015rethinking}   &     78.4 $\pm$ 0.2 \comment{33gp54sr68}  &  74.6 $\pm$ 0.1 \comment{4bjiewv89m}     \\
        \midrule
        Ours (Instance)             &        \textbf{79.2 $\pm$ 0.2} \comment{S2 7zww5zc8bi S3 c9w2rags7c S1 ifxp44k6sq} & \textbf{75.3 $\pm$ 0.3} \comment{3acmxmvd8b S2 d5dgdh9fbb S1 igfz37r7xq}  \\ \bottomrule
        \end{tabular}
    }
  \captionof{table}{On CIFAR100, our method of learning dynamic labels outperform common methods that use fixed labels.}
  \label{tab:other-baselines}

\end{wrapfigure}
\vspace{-0.25cm}
\noindent\textbf{Comparison with other methods}: In Table \ref{tab:other-baselines}, we compare our approach to methods that use fixed labels, but incorporate other strategies for improving generalization.
Confidence penalty \cite{pereyra2017regularizing} uses one-hot labels to optimize the model parameters, but regularize the model predictions by penalizing low entropy distributions. Label smoothing \cite{szegedy2015rethinking} regularizes over-confident predictions by optimizing soft labels, that remain fixed throughout optimization. These soft labels are obtained by modifying the one-hot labels to allocate a small amount of uniform weights to non target classes. In contrast, our method takes into account the shared concepts at different stages of training, and dynamically regularizes the model for softer decision boundaries. For similar reasons stated above, we obtain diminished gains on CIFAR10 (reported in Supplementary).

\subsection{Analysis of dynamic labels}
\label{sec:experiments-analysis}
In the previous section, we establish that dynamic soft labels improve generalization. In this section, we study the different components of our method and highlight their importance. 

\begin{wrapfigure}{r}{0.40\columnwidth}
  \centering
    \resizebox{0.30\columnwidth}{!}{
    \begin{tabular}{@{}ccc@{}}
        \toprule
        Dynamic  & Class & Instance \\ \midrule
        \xmark   &   77.2 $\pm$ 0.1 \comment{y4tkcz2fv4} & 78.4 $\pm$ 0.3 \comment{3zdhy4kdze}     
        \\
        \cmark         &  \textbf{78.8 $\pm$ 0.2} \comment{m5qgy9rzfp} & \textbf{79.2 $\pm$ 0.2}  \comment{5grp7w33ii}
        \\
        \bottomrule
        \end{tabular}
    }
  \captionof{table}{Comparison of ResNet18 on CIFAR100 with dynamic and static soft labels.}
  \label{tab:dynamic-vs-static}
\vspace{-0.1in}
\end{wrapfigure}
\noindent\textbf{Importance of dynamic labels}: To study the importance of trajectories of labels for generalization, we train a model in two settings (1) dynamic label trajectories {updated} at each iteration, and (2) {fixed} labels {taken from setting (1)} at convergence. As shown in Figure \ref{tab:dynamic-vs-static}, the model trained with {fixed} labels at convergence performs worse. This result empirically establishes that we need different degrees of smoothing (hence, dynamic labels) throughout optimization to improve generalization.

\begin{figure}[!htbp]
\centering
\begin{minipage}[b]{0.36\textwidth}
  \centering
    \includegraphics[scale=0.20]{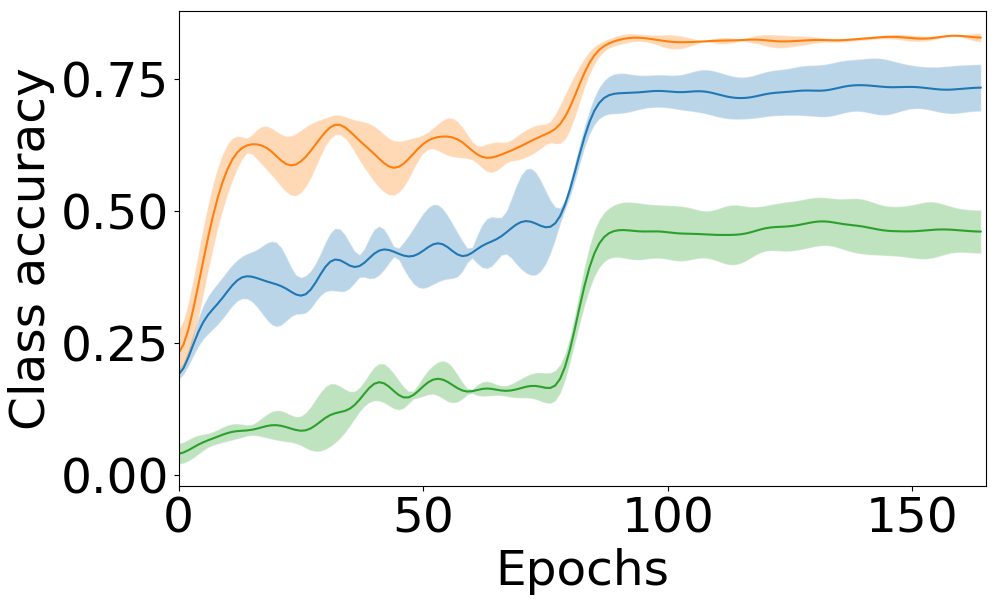}
\end{minipage}%
\hfill
\begin{minipage}[b]{0.36\textwidth}
  \centering
    \includegraphics[scale=0.20]{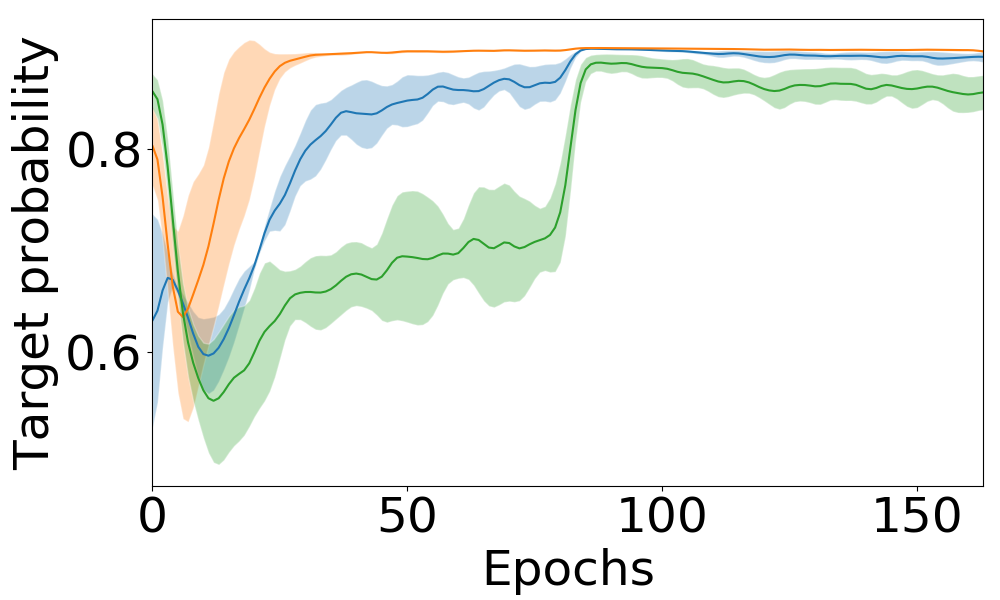}
\end{minipage}%
\hfill
\begin{minipage}[b]{0.28\textwidth}
  \centering
    \includegraphics[scale=0.30]{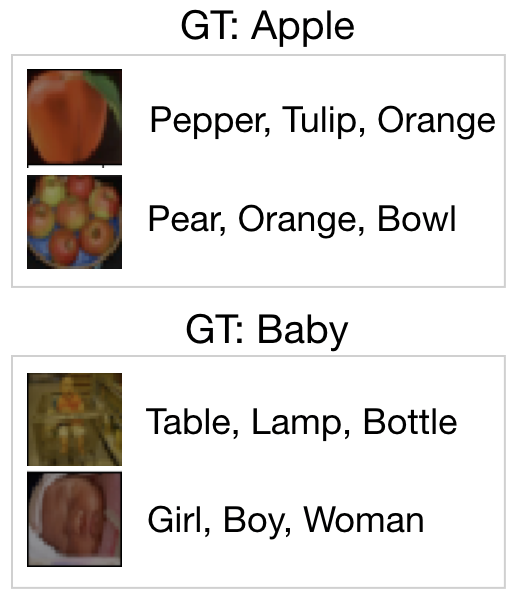}
\end{minipage}%
\caption{Learning dynamics of soft labels on ResNet18 for CIFAR100 across three classes \textbf{Left}: Meta dataset accuracy of three random classes during optimization \textbf{Middle}: Regularization trajectory of these classes, where harder classes are regularized more at the start of optimization. \textbf{Right}: Top three non-target classes in instance labels capture the semantic properties of the instance. For example \textit{Bowl} being a class for an instance where \textit{Apples} are placed in a bowl. \textit{Baby} had \textit{Girl, Boy} and \textit{Woman} as top classes because they are semantically similar to the target class.}
\label{fig:dynamic_global_trajectory}
\end{figure}

\noindent\textbf{Dynamic labels are repeatable}:
To perform a qualitative evaluation of the trajectories, we visualize the class labels in Figure \ref{fig:dynamic_global_trajectory} (middle). For CIFAR100 using ResNet18, we pick three random classes, and plot the probability mass on target class over the course of training (mean and standard deviation over three runs). The trajectories are dynamic, and adapt differently for these classes. More importantly, low standard deviation implies the learned trajectories are highly repeatable, and intrinsic to the dataset and the model. 

\noindent\textbf{Dynamic labels provide regularization}:
In Figure \ref{fig:dynamic_global_trajectory} (left), we plot the meta set accuracy of three classes during optimization, which aligns with our hypothesis that certain classes are harder to optimize, and hence, need to be regularized differently. In Figure \ref{fig:dynamic_global_trajectory} (middle), we see that the target class probability (i.e. the effective smoothing due to the class label) and its meta set accuracy are highly correlated. E.g. the class depicted in green has the lowest accuracy, and therefore the highest smoothing. As the model performance improves, the smoothing reduces. This dynamic behavior is important for generalization, but is not possible in common methods which optimize model parameters with fixed labels \cite{pereyra2017regularizing, szegedy2015rethinking} or a fixed degree of regularization \cite{srivastava2014dropout}. 

\noindent\textbf{Instance labels capture instance properties}: 
Instance labels relax the assumption that non-target classes are equally likely. Therefore, this method can learn {more detailed} softer decision boundaries, which helps to capture instance specific properties. As shown in Figure \ref{fig:dynamic_global_trajectory} (right), our method learned different instance labels for two samples of the same class \emph{Baby}. For example, in the image of a \textit{baby} sitting on a \textit{table}, the second-most probability mass is on class \textit{Table}. Similarly, the learned soft labels also account for similar concepts between classes. For example, in the images of an \textit{apple}, the most probable classes are similar to target class i.e. \textit{pear} and \textit{orange}.

\vspace{-0.25cm}

\subsection{Improving knowledge distillation using instance labels}

 Distillation \cite{hinton2015distilling} is a method to transfer the generalization ability of a large (teacher) model into a compact (student) model, via the information present in the predictions of the teacher model. More precisely, the predictions from a pretrained teacher model are softened using a temperature parameter, and used as targets for training the student model. Intuitively, methods such as label smoothing which regularize the models against making over-confident predictions, should help to improve the teacher models, thereby improving the downstream task of distillation. However, recent work \cite{Rafael2019} has shown that training teacher models with label smoothing leads to inferior student models, as the instance specific information is lost. In contrast, in this section, we show that training teacher models with dynamic labels makes them more suitable for distillation. 

\begin{table}[!htbp]
\centering
\small\begin{tabular}{ccccc}
\toprule
Temperature scaling       & Baseline & Label Smoothing   & Ours (Model) & Ours (Labels) \\ \midrule
  \xmark  &     $42.1 \pm 0.6$ \comment{wsigi9fqn8}     &  $43.2 \pm 0.1$ \comment{wsigi9fqn8}   &  $\bf{45.2 \pm 0.3}$ \comment{9yiktqz32u}   &    $45.1 \pm 0.8$ \comment{tti5m76qc2}   \\
 \cmark & $45.2 \pm 0.7$  \comment{d6uf67spu3}    &  $44.2 \pm 0.4$ \comment{uyhgjiv6py} &  $\bf{45.7 \pm 0.3}$ \comment{8iydptf3nq}    & $\bf{45.7 \pm 0.2}$  \comment{my9ftnmnw5}       \\ \bottomrule
\end{tabular}
\caption{Performance of distillation from ResNet18 to AlexNet on CIFAR100 using instance labels.}
  \label{tab:distillation}
\end{table}

\vspace{-0.25cm}
For this study, we replicate the setting in \cite{Rafael2019}, and use ResNet18 as teacher model and AlexNet as student model. We compare with knowledge distillation using one-hot targets and label smoothing. We use instance labels in two settings (1) Similar to \cite{hinton2015distilling}, we use predictions (referred as \textit{Ours (Model)} in Table \ref{tab:distillation}) from a teacher model trained with instance labels, as targets for the student, and (2) we use the converged instance labels (referred as \textit{Ours (Labels)} in Table \ref{tab:distillation}) as targets for the student. In all experiments, the temperature is tuned by searching in the range of [0, 10] with increments of 0.25. 

First, we replicate the findings of \cite{Rafael2019} in Table \ref{tab:distillation}, and verify that training with label-smoothing leads to worse downstream performance for distillation. As shown in \cite{Rafael2019}, use of a uniform-prior in label smoothing pushes all classes equally far, destroying the semantic similarity between classes and instances. In contrast, our method treats classes heterogeneously, and learns label smoothing specific to an instance (see Figure \ref{fig:dynamic_global_trajectory} (right)). Accordingly, as shown in Table \ref{tab:distillation}, when we distill knowledge from a teacher model trained using our method, we outperform the teacher trained using label smoothing by a significant margin. Our teacher model trained with instance labels is also less sensitive to temperature tuning, and performs well even without temperature scaling. This is expected as unlike distillation with one-hot targets where the predictions are softened at convergence, we optimize using soft labels at each step of optimization.

Now that we have established the benefit of training teacher models with instance labels, we evaluate the efficacy of instance labels obtained at convergence, for the purpose of distillation. As shown in Table \ref{tab:distillation}, we achieve comparable performance even when we use soft labels instead of model predictions as targets for distillation. This shows that the converged labels capture the \textit{dark knowledge} \cite{hinton2015distilling} and can be used for label distillation. We expect this performance to further improve by using dynamic label trajectories, instead of static labels at convergence. However, due to difference in factors like learning rate schedule, and number of iterations, this is not straight-forward. We leave this for future exploration.

\begin{wrapfigure}{r}{0.40\columnwidth}
  \centering
  \small
    \resizebox{0.40\columnwidth}{!}{
  \begin{tabular}{@{}ll|l|l@{}}
        \toprule
        Train/Transfer  & VGG16         & ResNet18 & AlexNet \\ \midrule
        VGG16           &  $76.5 \pm 0.2$  \comment{5cn38vs2ya} & $\bf{79.6 \pm 0.2}$ \comment{8ad2ijp9pd} & $45.1\pm 0.2$ \comment{v7qfzbfaj9} \\
        ResNet18        &  $\bf{76.9 \pm 0.3}$ \comment{bsphs83jae} &  $79.1 \pm 0.3$  \comment{wcw3fie4qk} & $\bf{45.7 \pm 0.2}$ \comment{my9ftnmnw5}\\ 
        \midrule
        Baseline     & 74.5 $\pm$ 0.2  &   77.1  $\pm$ 0.3 & 42.0 $\pm$ 0.4 \comment{myrn7hv6j9} \\ 
        \bottomrule
        \end{tabular}
    }
  \captionof{table}{Transferring converged instance labels for CIFAR100 across architectures.}
  \label{tab:transfer_soft_labels}
\vspace{-0.25cm}
\end{wrapfigure}

\noindent\textbf{Instance labels are generalizable across architectures}: 
From the above results, we would expect that instance labels depend more on the underlying data than on the architecture they were trained on. In this section, we show that instance labels {from one architecture} can be used to train other architectures. More specifically, we use the labels from VGG16 and ResNet18 at convergence, perform temperature scaling on them, and use them to train other architectures. As observed in Table \ref{tab:transfer_soft_labels}, using labels from other architectures outperforms the baseline by a significant margin. Interestingly, labels trained with ResNet18 generally perform better compared to labels trained using VGG16. We attribute this to the capacity of the model, which {we suppose could} play a crucial role in learning better labels.


\subsection{Image classification in a noisy label setting }

\label{sec:noisy}
\begin{wrapfigure}{r}{0.40\columnwidth}
  \centering
  \small
    \resizebox{0.40\columnwidth}{!}{
        \includegraphics[]{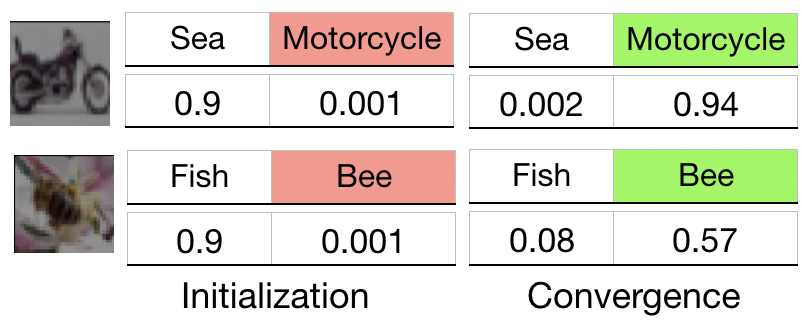}
    }
  \caption{Examples of noisy instances in CIFAR100 that were corrected using our framework.}
  \label{fig:correction of labels}
\vspace{-0.1in}
\end{wrapfigure}

As mentioned earlier, an ideal framework to learn from noisy data should not only ignore incorrectly labelled instances, but also be able to correct labels so as to learn from them. As shown in Figure \ref{fig:correction of labels}, instance labels shift the probability mass across classes during the course of optimization, thereby correcting noisy instances. In this section, we show how our method performs in a noisy label setting by correcting incorrect samples, and how this compares with the existing state-of-art. 

To create the noisy data setting, and to compare with the relevant state-of-art, we follow the common settings as mentioned in \cite{jiang2017mentornet, saxena2019data, shu2019metaweightnet}. Here, the target label of each instance is independently changed to a uniformly random class with probability $p$. This $p$ is {called} the \emph{noise fraction}, and we show results with $p$=[0.2, 0.4, 0.6 and 0.8]. The target labels in the validation and meta set are, however, not changed. We maintain the same setting in \cite{jiang2017mentornet, shu2019metaweightnet} and use WideResNet-28-10 architecture, and 2\% training set as clean meta data. We report the results on CIFAR100 and CIFAR10 in two settings - (1) Setting A as in \cite{shu2019metaweightnet}, and (2) Setting B as in \cite{saxena2016convolutional}.

\begin{wrapfigure}{r}{0.40\columnwidth}
  \centering
  \small
    \resizebox{0.34\columnwidth}{!}{
        \includegraphics[]{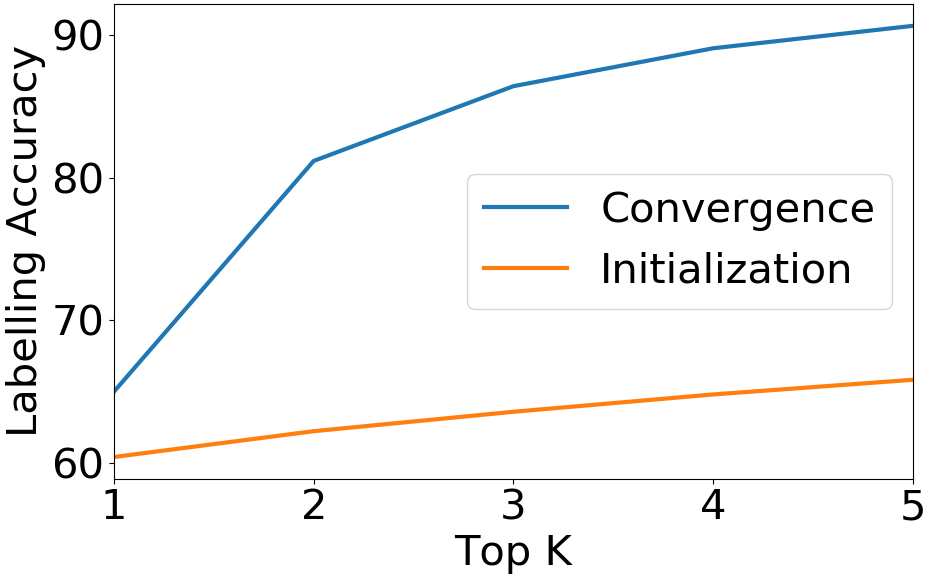}
    }
  \caption{Instance labels correct noise in CIFAR100 by shifting the probability mass on the correct class during training. Top-5 accuracy changes from 65\% at start of training to 90\% at convergence.}
  \label{fig:topk}
\vspace{-0.1in}
\end{wrapfigure}

Current state-of-art approaches assign a weight to each instance in the training set, where ideally the noisy instances should be down-weighted over the course of training (see Section \ref{sec:related-work}. However, instance labels have an additional advantage in that they are able to change the target labels, there-by correcting the noisy instances. For example, at convergence, our method corrected 7\% of the noisy samples of CIFAR100 in the 40\% noise setting (see Figure \ref{fig:topk}), reaching 33\% noise at top-1 at convergence. Further, we observe that this method reduces the entropy of the noisy samples which it cannot correct, by reducing the probability mass on the incorrect target classes. We leverage this property to learn an implicit curriculum over the instances, by weighing each instance by the entropy of its label. In setting A \cite{shu2019metaweightnet}, our method outperforms the current state-of-art in CIFAR10 by 1.1\% at 40\% noise, and by 1.6\% at 60\% noise. On CIFAR100, we outperform the state-of-art by 2\% at 40\% noise, and 0.9\% at 60\% noise (see Table \ref{tab:mwnet-noisy}). Finally, in setting B \cite{saxena2019data}, we outperform state-of-art in CIFAR100 by 1.2\% at 20\% noise, and 0.9\% at 40\% noise. In the extreme noise setting at 80\%, our method achieves comparable performance to the current state of art (see Table \ref{tab:mwnet-noisy_2}).

\begin{minipage}{\textwidth}
\begin{minipage}[b]{0.48\textwidth}
    \centering
\resizebox{1.0\textwidth}{!}{%
  \begin{tabular}{lcc}
    \toprule
                                               &      40\%            &    60\%        \\ 
    \midrule
    Baseline  \cite{shu2019metaweightnet}               & $68.1 \pm 0.2$       & $53.1 \pm 3.0$  \\ 
    Focal Loss \cite{lin2017focal}             & $76.0 \pm 1.3$       & $51.9 \pm 1.2$  \\ 
    Co-teaching \cite{han2018co}               & $74.8 \pm 0.3$     & $73.1 \pm 0.3$        \\ 
    \midrule
    \multicolumn{3}{c}{Using Additional Clean Data}\\
    \midrule
    MentorNet   \cite{jiang2017mentornet}      & $87.3 \pm 0.2$       & $82.8 \pm 1.4$         \\ 
    L2RW        \cite{ren2018learning}         & $86.9 \pm 0.2$       & $82.2 \pm 0.4$        \\ 
    MWNet       \cite{shu2019metaweightnet}             & $89.3  \pm 0.3$      & $84.1 \pm 0.3$        \\ 
    Ours (Instance labels)                      & $\bf{90.4  \pm 0.1}$ & $\bf{85.7 \pm 0.2}$       \\ 
    \bottomrule
  \end{tabular}
}
   
\end{minipage}
\hfill
\begin{minipage}[b]{0.48\textwidth}
    \centering
\resizebox{1.0\textwidth}{!}{%
  \begin{tabular}{lcc}
    \toprule
                                               &      40\%            &    60\%        \\ 
    \midrule
    Baseline  \cite{shu2019metaweightnet}               & $51.1 \pm 0.4$       & $30.9 \pm 0.3$  \\ 
    Focal Loss \cite{lin2017focal}             & $51.2 \pm 0.5$       & $27.7 \pm 3.7$  \\ 
    Co-teaching \cite{han2018co}               & $46.2 \pm 0.15$     & $35.7 \pm 1.2$        \\ 
    \midrule
    \multicolumn{3}{c}{Using Additional Clean Data}\\
    \midrule
    MentorNet   \cite{jiang2017mentornet}      & $61.4 \pm 4.0$       & $36.9 \pm 1.5$         \\ 
    L2RW        \cite{ren2018learning}         & $60.8 \pm 0.9$       & $48.2 \pm 0.3$        \\ 
    MWNet       \cite{shu2019metaweightnet}             & $67.7  \pm 0.3$      & $58.8 \pm 0.1$        \\ 
    Ours  (Instance labels)                           & $\bf{69.7  \pm 0.3}$ & $\bf{59.4 \pm 0.4}$       \\ 
    \bottomrule
  \end{tabular}
}
    
\end{minipage}

\captionof{table}{Learning soft labels with WRN28-10 under varying uniform noise, outperforms the state-of-the-art methods. Model is trained with setting A \cite{shu2019metaweightnet} \textbf{Left}: CIFAR10 \textbf{Right}: \cite{shu2019metaweightnet}: CIFAR100}
\label{tab:mwnet-noisy}
\end{minipage}

\begin{table}[!htbp]
\centering
\small\begin{tabular}{lccc}
    \toprule
                                               &   20\% & 40\%     & 80\%   \\
    \midrule
    Baseline  \cite{ren2018learning}           &  $60.0$  &  $50.66 \pm 0.24$       &       $8.0$        \\
    MentorNet PD \cite{jiang2017mentornet}     &  $72.0$ &    $56.9$           &      $14.0$              \\
    Data Parameters \cite{saxena2019data}      &  $75.68 \pm 0.12$ & ${70.93 \pm 0.15}$  &  $\bf{35.8 \pm 1.0}$  \\
    \midrule
    \multicolumn{4}{c}{Using Additional Clean Data}\\
    \midrule
    MentorNet DD \cite{jiang2017mentornet}     & 73.0   &     $67.5$         &    $35.0$          \\
    L2RW         \cite{ren2018learning}        & - & $61.34 \pm 2.06$       &      -              \\
    Ours  (Instance labels)                            & $\bf{76.9 \pm 0.3}$ & $\bf{71.8 \pm 0.2}$        &   $\bf{35.6 \pm 0.6}$                 \\
    \bottomrule
  \end{tabular}
\caption{Learning soft labels with WRN28-10 under varying uniform noise, outperforms the state-of-the-art methods. Model is trained with setting B \cite{saxena2019data}: CIFAR100}
  \label{tab:mwnet-noisy_2}
\end{table}

\section{Related Work}
\label{sec:related-work}

\noindent\textbf{Soft labels}: Various techniques have been proposed to regularize model parameters for improving generalization \cite{ioffe2015batch, pereyra2017regularizing, srivastava2014dropout}. A relevant work is label smoothing \cite{szegedy2015rethinking}, where instead of the model parameters, the one-hot labels are regularized to a weighted mixture of targets in the dataset. This technique has been widely adopted in different problems like image-classification \cite{real2019regularized, zoph2018learning}, reducing word-error rate \cite{chorowski2016towards}, improving machine translation \cite{vaswani2017attention} etc. A shortcoming of this approach is that soft-labels remain static throughout the course of optimization. Multiple works \cite{dogan2019label,li2020regularization, pham2020meta, xie2016disturblabel} have been proposed to mitigate the static nature of labels. DisturbLabel \cite{xie2016disturblabel} regularizes model training by performing label dropout. Label smoothing can be viewed as a marginalized form of label dropout. \cite{dogan2019label} embed labels using word embeddings in a Euclidean space, and computes similarities (soft-labels) based on dot-product in this space. Their method is limited to settings where labels are comprised of natural words. \cite{li2020regularization} argue against using a single smoothing across entire dataset, and instead cluster the data and propose a heuristic to set the smoothing parameter for data points in each cluster. Our work does not rely on clustering in a feature space, and learns optimal soft-labels for each instance via meta-learning. \cite{li2019learning} also use meta-learning to learn optimal soft labels, but reset the label estimates at every epoch. On the contrary, our work treats labels as learnable parameters, that are continuously optimized throughout training (which is key to label correction in corrupt data setting). \cite{pham2020meta} is the most relevant to our proposed approach, where an auxillary teacher model is trained to assign soft targets to a student, thereby yielding dynamic soft labels. Our work does not rely on a teacher model, instead it explicitly instantiates and learns soft labels for the dataset. As a result, our method can account for the class membership of data points, and learns to regularize each class and instance separately. Also, their method is Markovian in nature, while our method benefits from the past history of optimization of the soft-labels.

\noindent\textbf{Noisy data setting}:
In the noisy data setting, \cite{jiang2017mentornet, shu2019metaweightnet} trained an auxillary neural network using meta-learning to assign weights 
to samples in the mini-batch.  \cite{ren2018learning} uses a held-out set to perform an online approximation of weights for samples in a minibatch. \cite{saxena2019data} introduced learnable temperature parameters per data-point, which scale the gradient contribution of each data-point. 
\cite{zhang2020distilling} use meta-learning to perform a linear combination of the original labels and pseudo labels generated by DNN on the fly. The learned weights are used to cluster the data as clean or mislabelled. While these works can reduce the contribution of noisy instances, they cannot correct the incorrect labels. In contrast, our framework is able to correct the labels of noisy instances, and outperforms these methods. 
\cite{tanaka2018joint, yi2019probabilistic} propose a framework for correcting noisy annotations in the dataset. 
They update model parameters and labels in an alternate manner using the training dataset. 
However, since the model parameters and labels are updated on the same data, they rely on hand-crafted heuristics to prevent degenerate solutions.
While they obtain promising results in low noise setting, these heuristics fail to converge in presence of high annotation noise \cite{yi2019probabilistic}. Our method does not rely on any heuristic, instead we use the meta-learning constraint, and outperform state-of-the-art at all levels of noise.
 
\section{Conclusion}
In this work, we have proposed a meta-learning framework under which the labels are treated as learnable parameters, and are optimized along with model parameters.  The learned labels take into account the model state, and provide dynamic regularization, thereby improving generalization.
We learn two categories of soft labels, class labels specific to each class, and instance labels specific to each instance. In case of supervised learning, training with dynamically learned labels leads to improvements across different datasets and architectures. In presence of noisy annotations in the dataset, our framework corrects annotation errors, and improves over the state-of-the-art. Finally, we show that teacher models trained with dynamically learned labels improves the downstream task of distillation.

\newpage
 {\small
 \bibliographystyle{ieee_fullname}
 \bibliography{egbib}
 }

\appendix
 
\newpage
\section*{Appendix}
\noindent\textbf{Comparison of different methods on CIFAR10}

\begin{table}[!htbp]
\small
\centering
\begin{tabular}{@{}lll@{}}
\toprule
                     
\multicolumn{1}{c}{} & \multicolumn{1}{c}{ResNet18} & \multicolumn{1}{c}{VGG16} \\ \midrule
Label Smoothing             &    95.1 $\pm$ 0.3 & 94.2 $\pm$ 0.3                               \\
Confidence Penalty & 95.2 $\pm$ 0.1 & 94.0 $\pm$ 0.2 \\\midrule
Ours (Class)               &  95.2 $\pm$ 0.1 \comment{z6evrs493t zec43tsuj9 8by6q24ga8} & 94.1 $\pm$ 0.1 \comment{adu6wntpt4 4y88ikyn9n bzgd33nnys} \\
Ours (Instance)            &  \textbf{95.3 $\pm$ 0.1}  \comment{mvxfv33ij4 gqkac9zqa7 qb3q3qepxj} & \textbf{94.3 $\pm$ 0.1} \comment{hkdpb7fftk gzd442byam v84hf6e6df} \\ \bottomrule
\end{tabular}
\caption{On CIFAR10, our method of learning dynamic soft labels is comparable to common methods that use fixed labels}
\label{tab:baseline_results}
\end{table}

\noindent\textbf{Hyperparameters for CNN architectures}
\begin{table}[!htbp]
\small
\centering
\begin{tabular}{@{}llll@{}}
\toprule
              & ResNet18 & VGG16 & WRN28-10 \\ \midrule
Learning rate & 0.4      & 0.6 &  0.5 \\
Weight decay  & 1e-3     & 5e-3  & 1e-3   \\
Epochs  & 165      & 300 &  120 \\ \bottomrule
\end{tabular}
\caption{Hyperparameter setting for CNN architectures. We do not use dropout.}
\end{table}

\noindent\textbf{Hyperparameters for dynamic soft labels }

The hyperprameters are tuned on a single cross validation fold using 20\% training data as held-out data.  
For class labels, we use a learning rate search grid of [1e-2, 3e-2, 1e-1, 3e-1, 1, 3]. For instance labels, we use a learning rate search grid of [5, 10, 25, 50, 75, 100]. For initialization of soft labels, we use a search grid of [0.3, 0.5, 0.7, 0.9] for both, class and instance labels. This initialization value is used to initialize the target class probability. The non-target class is initialized with a uniform distribution using the remaining probability.

\newpage

\end{document}